\documentclass[letterpaper]{article} 
\usepackage[draft]{aaai2026}  
\usepackage{times}  
\usepackage{helvet}  
\usepackage{courier}  
\usepackage[hyphens]{url}  
\usepackage{graphicx} 
\usepackage{natbib}  
\usepackage{caption} 
\usepackage{algorithm}
\usepackage{algorithmic}

\usepackage{preamble}
\usepackage{newfloat}
\usepackage{listings}
\DeclareCaptionStyle{ruled}{labelfont=normalfont,labelsep=colon,strut=off} 
\floatstyle{ruled}
\newfloat{listing}{tb}{lst}{}
\floatname{listing}{Listing}
\title{Reconstruction Using the Invisible: Intuition from NIR and Metadata \\ 
        for Enhanced 3D Gaussian Splatting}
\author{
  Gyusam Chang$^{1,2}$\thanks{Work done while the author was a visiting researcher at UCLA.}
  \qquad Tuan-Anh Vu$^{2}$
  \qquad Vivek Alumootil$^{2}$ \\
  \qquad Harris Song$^{2}$ 
  \qquad Deanna Pham$^{2}$
  \qquad Sangpil Kim$^{1}$$^\dagger$ 
  \qquad M. Khalid Jawed$^{2}$\thanks{Corresponding authors.} \\
}
  
\affiliations{
    \textsuperscript{\rm 1}Korea University 
    \qquad\textsuperscript{\rm 2}University of California, Los Angeles\\
    \{gsjang95, spk7\}@korea.ac.kr \qquad \{tuananh.vu, vivekalumootil, songharris2006, deannapham2004, khalidjm\}@ucla.edu

}

\begin{document}

\maketitle

\begin{abstract}

While 3D Gaussian Splatting (3DGS) has rapidly advanced, its application in agriculture remains underexplored. 
Agricultural scenes present unique challenges for 3D reconstruction methods, particularly due to uneven illumination, occlusions, and a limited field of view. 
To address these limitations, we introduce \textbf{NIRPlant}, a novel multimodal dataset encompassing Near-Infrared (NIR) imagery, RGB imagery, textual metadata, Depth, and LiDAR data collected under varied indoor and outdoor lighting conditions. 
By integrating NIR data, our approach enhances robustness and provides crucial botanical insights that extend beyond the visible spectrum. 
Additionally, we leverage text-based metadata derived from vegetation indices, such as NDVI, NDWI, and the chlorophyll index, which significantly enriches the contextual understanding of complex agricultural environments. 
To fully exploit these modalities, we propose \textbf{NIRSplat}, an effective multimodal Gaussian splatting architecture employing a cross-attention mechanism combined with 3D point-based positional encoding, providing robust geometric priors. 
Comprehensive experiments demonstrate that \textbf{NIRSplat} outperforms existing landmark methods, including 3DGS, CoR-GS, and InstantSplat, highlighting its effectiveness in challenging agricultural scenarios. The code and dataset are publicly available at: \url{https://github.com/StructuresComp/3D-Reconstruction-NIR}

\end{abstract}    
\section{Introduction}
\label{sec:intro}

3D reconstruction has become increasingly crucial across various fields, including robotics, autonomous driving, augmented reality, and agricultural monitoring. Traditional methods for reconstructing three-dimensional structures from two-dimensional images often struggle to capture fine details, handle complex scenes, and maintain robustness under challenging environmental conditions. Recently, 3D Gaussian Splatting (3DGS)~\cite{3dgs} has emerged as a significant advancement, enabling smoother, more detailed, and computationally efficient reconstructions. Unlike traditional approaches that rely heavily on discrete point representations~\cite{sinha2017surfnet, li2018efficient, lin2018learning, nguyen2019graphx}, 3DGS represents each 3D point as a Gaussian distribution, effectively capturing uncertainties and spatial continuity in complex environments.

\begin{figure}[t]
    \centering
    \includegraphics[width=0.8\linewidth]{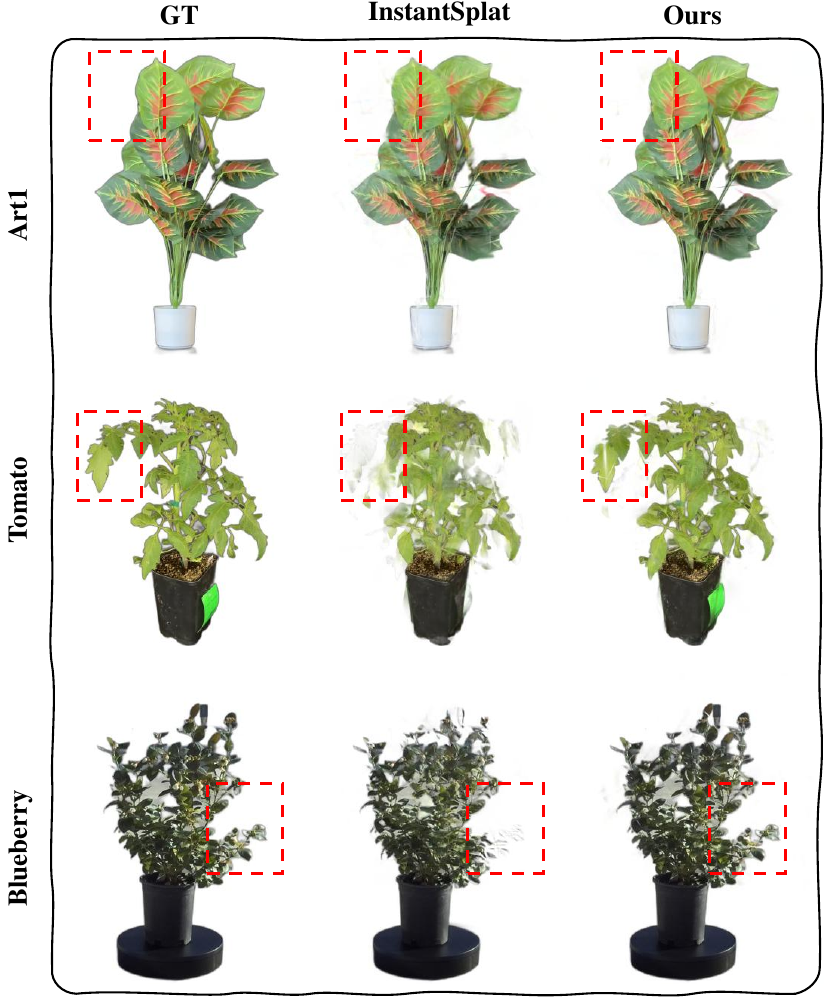}
    \vspace{-1em}
    \caption{\small Qualitative comparisons in a 3-view setup: rendered images from the ground truth, InstantSplat, and our method. We highlight our improved semantic understanding, particularly in regions of interest within the image, where our method more accurately captures meaningful structures and distinctions under novel lighting conditions. Zoom in for better visualization.
    }
    \label{fig:teaser}
    \vspace{-1em}
\end{figure}

Despite its demonstrated success in general scenarios, 3DGS faces significant challenges when applied to agriculture. As shown in Fig.~\ref{fig:teaser}, these environments pose unique challenges, including unpredictable lighting variations (\eg, intense sunlight, low visibility, and sunset conditions), limited viewing angles, environmental instability due to weather fluctuations, and frequent occlusions by foliage. These factors can substantially degrade the performance of typical 3D reconstruction methods~\cite{fan2024instantsplat, zhang2024cor}, resulting in incomplete or inaccurate plant modeling.

To overcome these limitations, we introduce the \textbf{NIRPlant} dataset, specifically designed to address the unique challenges of agricultural environments. \textbf{NIRPlant} incorporates comprehensive multimodal data, including RGB images, Near-Infrared (NIR) imagery, and rich textual metadata. 
The dataset (see \cref{tab:dataset_comparison}) comprises diverse indoor and outdoor lighting scenarios captured from multiple perspectives, including artificial illumination, direct sunlight, and sunset conditions. Integrating NIR imagery is particularly advantageous because NIR can capture plant-specific reflectance characteristics invisible to conventional RGB cameras, thus providing essential botanical information about plant health, water content, and structural integrity. For instance, high values of NDVI (Normalized Difference Vegetation Index) typically indicate robust vegetation health, NDWI (Normalized Difference Water Index) reflects water content, and chlorophyll index values directly correlate with photosynthetic efficiency and plant vigor. Such indices enrich our dataset and significantly enhance the model's ability to accurately interpret complex botanical scenarios, as shown in Fig.~\ref{fig:teaser}.

Moreover, textual metadata derived from both RGB and NIR images includes environmental conditions, precise lighting descriptions, and quantitative botanical indices, thus providing rich context for reconstructing detailed 3D models. Fusing this metadata with visual modalities enables our method to interpret plants more effectively and model them under diverse and challenging photometric conditions.

To leverage the full potential of our multimodal dataset, we propose \textbf{NIRSplat}, a novel Gaussian splatting framework optimized for multimodal data integration. \textbf{NIRSplat} employs a novel cross-attention mechanism~\cite{zhu2020deformable,vaswani2017attention} that effectively combines NIR embeddings with RGB features. Our approach achieves superior scene understanding by exploiting the complementary strengths of RGB imagery and NIR-derived features. Moreover, inspired by the success of Vision-Language Models (VLMs)~\cite{radford2021learning, li2022blip, li2023blip, liu2023visual}, we integrate textual embeddings derived from metadata descriptions to further enhance semantic understanding. This multimodal interaction is further enhanced by employing a novel 3D point-based positional encoding method, which leverages spatial coherence from geometric priors to align and enrich 2D image features with 3D spatial information.

We conducted extensive evaluations to validate our proposed method, comparing \textbf{NIRSplat}  with state-of-the-art approaches. Our results show that \textbf{NIRSplat}  outperforms existing methods in terms of reconstruction accuracy, robustness to varying environmental conditions, and visual quality. We provide detailed analyses that highlight the contributions of each modality to the overall improvement in performance.

In summary, our key contributions include:

\begin{itemize}
    \item The introduction of \textbf{NIRPlant}, a comprehensive multimodal agricultural dataset integrating RGB, NIR, and detailed textual metadata, enabling robust 3D reconstruction under varying lighting and environmental conditions.
    \item Development of \textbf{NIRSplat}, a multimodal Gaussian splatting framework employing cross-attention mechanisms and geometric priors, significantly improving scene reconstruction robustness.
    \item Extensive comparative analyses demonstrate our approach's effectiveness and advantages over leading methods, including 3DGS, CoR-GS, and InstantSplat, under diverse agricultural conditions (\eg, intense sunlight, occlusion, low visibility).
\end{itemize}

\section{Related Works}

\subsection{Method for 3D Reconstruction}

3D Gaussian Splatting (3DGS)~\cite{3dgs} uses a set of 3D Gaussian parameters and differentiable splatting to represent and render scenes more efficiently than traditional radiance fields~\cite{martin2021nerf, garbin2021fastnerf, barron2021mipnerf}.
Mip-Splat~\cite{Yu2024MipSplatting} is another scene rendering algorithm that constrains the size of 3D Gaussian primitives and mitigates aliasing and dilation issues present in 3DGS.
CF-3DGS~\cite{fu2024colmap} reduces the burden of pre-computation by leveraging the temporal continuity from video and the explicit point cloud representation.
CoR-GS~\cite{zhang2024cor} identifies and suppresses inaccurate reconstruction using Co-pruning, considers Gaussians, and Pseudo-view co-regularization.
Furthermore, another InstantSplat~\cite{fan2024instantsplat} is compatible with both the above methods and specializes in low-image-count representations through a neural network representation similar to that of NSR, utilizing a Gaussian Bundle Adjustment (GauBA).
SplatFields~\cite{mihajlovic2024splatfields} designs and regularizes splat features as the outputs of a corresponding implicit neural field.
Recently, CATSplat~\cite{roh2024catsplat} introduced a generalizable transformer-based framework, addressing the inherent constraints in monocular settings.

\begin{table}[t]
	\begin{center}
        \caption{\small Comparison with existing landmark dataset. R, D, N, S, L, and T denote RGB, Depth, NIR, Structured-Light Scanner (SLS), LiDAR, and Text, respectively. Lighting indicates whether there are various lighting conditions for supervision. 
        }
        \vspace{-0.9em}
	    \label{tab:dataset_comparison}
    	\resizebox{\linewidth}{!}{%
    	\begin{tabular}{lccccc}
            \toprule
            \textbf{Dataset} & \textbf{Modality} & \textbf{Lighting} & \textbf{$\#$ Scenes} & \textbf{$\#$ Views} & \textbf{Metadata} \\
            \midrule
            \citet{barron2022mip} & R & \ding{55} & 9 & 100-330 &  \ding{55} \\
            \citet{Knapitsch2017}  & R & \ding{55} & 14 & 4-17 & \ding{55} \\
            \citet{toschi2023relight} & R & \ding{51} & 20 & 2000 & \ding{55} \\ 
            \citet{voynov2023multi} & R,D,S & \ding{55} & 107 & 100 & \ding{55}\\
            \cmidrule{1-6}
            NIRPlant (Ours) & R,D,N,L,T & \ding{51} & 34 & 360 & \ding{51} \\
           \bottomrule
        \end{tabular}}
    \end{center}
    \vspace{-1.5em}
\end{table}

\subsection{Dataset for 3D Reconstruction} 

Various 3D Reconstruction datasets have focused on advancements in lighting recognition and multimodal approaches. Mip-NeRF360~\cite{barron2022mip} synthesizes realistic object views in the real world, while MVimgNet enhances 3D capture through video-based 3D-aware signals~\cite{yu2023mvimgnet}. Adding on, ReLight and Tanks are designed to address lighting variation with different materials~\cite{toschi2023relight,voynov2023multi,Knapitsch2017}. Previous adaptations of NeRF to real-world environments through LEGO bricks~\cite{li2023mobilebrick} and famous city sites~\cite{martin2021nerf} improve lighting capture by training from photo datasets. OmniObject3D addresses surface reconstruction for dense and sparse-view surfaces~\cite{Wu_2023_CVPR}. Additionally, GauU-Scene~\cite{xiong2024gauu} supports large-scale scene reconstruction using Gaussian Splatting for real-time scanning. NeRFBK~\cite{yan2023nerfbk} utilizes both real and synthetic data to capture objects of varying materials and lighting conditions, thereby comparing NeRF in outdoor views and for transparent objects. UniSDF~\cite{NEURIPS2024_05b12f10} is another dataset that utilizes NeRF to capture 3D scenes with reflections, combining the traditional SDF with radiance fields to render scenes with and without reflections.

\section{Our Multimodal NIRPlant Dataset}
\label{sec:dataset}

\begin{figure}[t]
    \centering
    \includegraphics[width=1\linewidth]{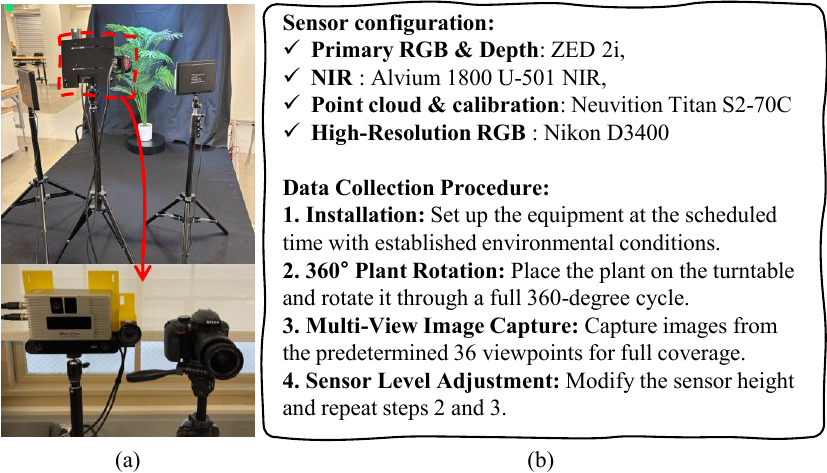}
    \vspace{-1.5em}
    \caption{\small (a) Data Acquisition Platform: (Top) Sensor installation, and (Bottom) Sensor configuration. (b) Sensor configuration and data collection procedure for both indoor and outdoor settings.}
    \label{fig:sensor}
    \vspace{-0.7em}
\end{figure}

Collecting comprehensive multimodal datasets in diverse environmental conditions is inherently challenging, even within controlled laboratory settings. The manual process involved in collecting, observing, and annotating multiple data types, such as RGB and NIR imagery, and metadata, is labor-intensive, time-consuming, and costly. Moreover, achieving diversity in visual data and ensuring high-quality annotations significantly complicates the process. Please refer to the supplementary material for additional details.

\subsection{Data Acquisition Platform}
\label{subsec:platform}

Our primary objective is to develop a comprehensive and versatile 3D plant reconstruction dataset under diverse real-world lighting conditions. However, dynamic environmental factors such as wind, shadows, and fluctuating sunlight pose significant challenges to data reliability. To mitigate these effects, we designed a controlled lighting configuration (see Fig.~\ref{fig:sensor}) that captures data during critical illumination periods, allowing us to observe lighting variance systematically. Precisely, objects were positioned to capture precise lighting conditions at defined times of the day (\ie, noon and sunset). Furthermore, we utilized a multimodal sensor setup consisting of a ZED 2i and Nikon D3400 HD RGB cameras, an Alvium 1800 U-501 NIR sensor, and a Neuvition Titan S2-70C LiDAR sensor, ensuring accurate alignment and consistent distance between the objects and sensors. To enhance data quality under natural conditions, automatic adjustments for focus, exposure, and gain were employed to maintain consistency and adaptivity in capturing agricultural data, thus enabling accurate calibration and precise extraction of camera perspectives essential for reliable 3D reconstruction. 

\begin{figure}[t]
    \centering
    \includegraphics[width=0.85\linewidth]{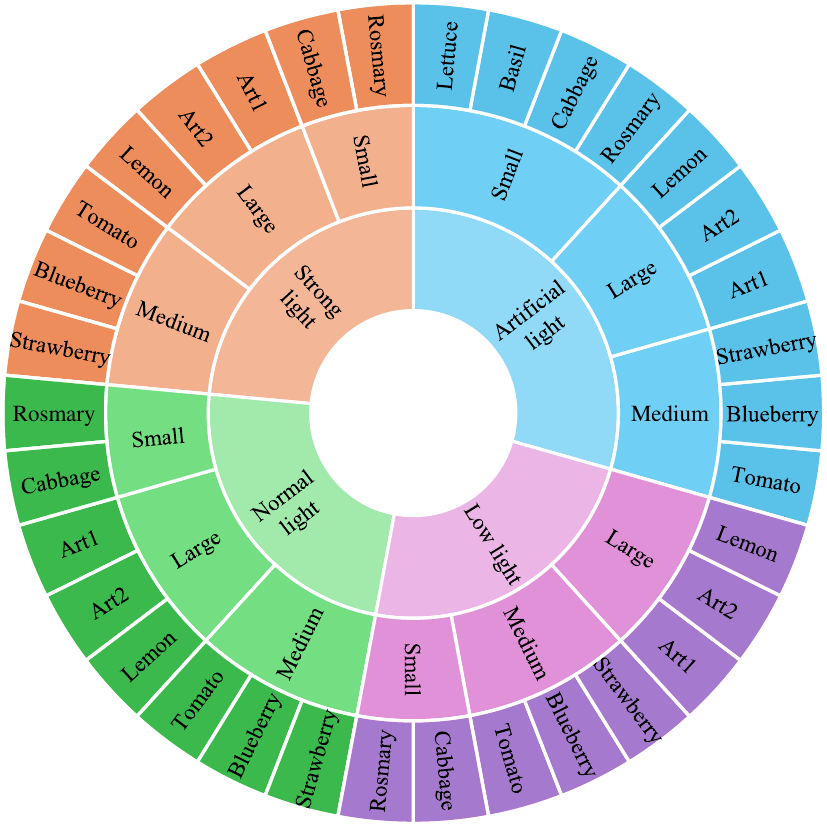}
    \vspace{-0.8em}
    \caption{Hierarchical organization of the dataset taxonomy.}
    \label{fig:taxonomic}
    \vspace{-0.8em}
\end{figure}

\subsection{Data Construction and Processing}
\label{subsec:construction}

Collecting precise camera poses in agricultural environments is inherently challenging due to the dynamic nature of plants, whose structures change rapidly in response to environmental factors. To overcome this issue, extensive data were collected indoors and outdoors under strictly consistent conditions. Specifically, each object was captured from 360 multi-modal data samples per scene, and ground truth models were constructed using the landmark Structure-from-Motion (SfM) technique~\cite{schoenberger2016sfm}, ensuring high accuracy and robustness, particularly for texture-rich environments. Additionally, precise reconstruction was prioritized by meticulously removing background information from object images. In total, our dataset contains comprehensive multimodal data (as described in Tab.~\ref{tab:dataset_comparison}), covering four distinct lighting scenarios with 360 viewpoints per scene across up to 10 different plant categories, ensuring broad diversity and representativeness (see Fig.~\ref{fig:taxonomic}).

\begin{figure*}
    \centering
    \includegraphics[width=0.95\linewidth]{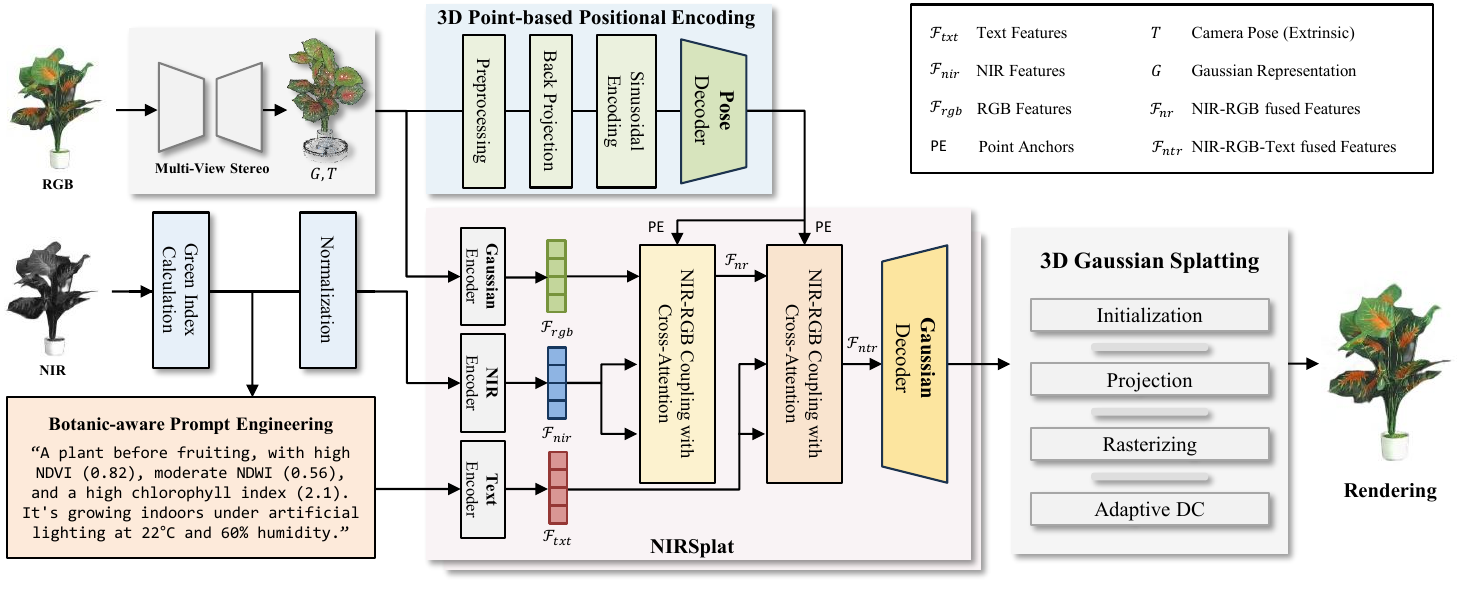}
    \vspace{-0.8em}
    \caption{The overall architecture of \textbf{NIRSplat} framework. NIRSplat efficiently processes tri-modal inputs consisting of NIR, RGB, and Text, enabling joint reasoning. The details of the prompt engineering are included in the supplementary document.}
    \label{fig:overall}
    \vspace{-0.8em}
\end{figure*}

\subsection{Dataset Specifications and Statistics}
\label{subsec:stat}

\myparagraph{Data Organization.}
As illustrated in  Fig.~\ref{fig:taxonomic}, our \textbf{NIRPlant} dataset categorizes plant data under four primary lighting conditions: artificial light, strong sunlight, low light, and normal daylight. It encompasses up to 10 diverse plant species, further classified into three distinct size categories (small, medium, and large). Each plant category and lighting condition was captured consistently from 360 viewpoints to ensure robust perspective coverage. This systematic data acquisition process was uniformly applied across RGB, NIR, Depth, and LiDAR. Note that botanic-aware prompts are generated for each scene, as illustrated in the supplementary material.
Additionally, to effectively extract discriminative NIR signals, comparisons were conducted against artificial plants (Art 1 and Art 2). Leveraging this comprehensive multimodal dataset structure, we aim to enhance the understanding of plant-specific characteristics under various environments, thus substantially improving the performance and robustness of 3D reconstruction methods.

\myparagraph{Dataset Split.}
Considering the practical agricultural environment, we adopted a sparse-view approach inspired by InstantSplat~\cite{instantsplat}. Specifically, from each set of 24 RGB, NIR, and textual metadata viewpoints, we randomly sampled 3, 6, and 12 views for training and testing purposes, respectively. This sampling strategy simulates real-world scenarios where limited views are common, ensuring the dataset's applicability and the generalization of reconstruction algorithms in realistic agricultural contexts.

\section{Our Proposed Method}
\label{sec:method}
In this section, we describe our multimodal method in detail, emphasizing the multimodal initialization through 3D positional encoding, the Transformer-based interactions for modality fusion, and the multimodal loss and regularization mechanisms. Note that the technical background necessary for understanding our proposed method is provided in the supplementary material.

\begin{table}[h!]
\centering
\begin{tabular}{lp{0.635\linewidth}}
\toprule
\textbf{Symbol} & \textbf{Description} \\
\midrule
$\text{G}$ & Set of learnable Gaussian primitives \\
$\{\bm{\mu}, \bm{\alpha}, \bm{\Sigma}, \bm{c}\}$ & Gaussian representations\\
$\text{P} = \{\mathbf{p}_i\}_{i=1}^N$ & 3D point maps \\
$\text{T} = [R \mid \mathbf{t}]$ & Camera pose \\
$\text{K}$ & Camera intrinsic matrix \\
$\mathbf{u}_i$ & 2D projection of point $\mathbf{p}_i$ via $T$ and $K$ \\
$\lambda$ & Learnable frequency scale \\
$\texttt{PE}_i$ & Positional encoding from projected 2D point \\
$\text{F}_{\text{rgb}}, \text{F}_{\text{nir}}, \text{F}_{\text{txt}}$ & Feature representations extracted from RGB images, NIR images, and text inputs, respectively \\
$\text{F}_{\text{nr}}$ & Joint visual representation obtained by fusing $\text{F}_{\text{rgb}}$ and $\text{F}_{\text{nir}}$ \\
$\text{F}_{\text{ntr}}$ & Multimodal feature representation obtained by integrating $\text{F}_{\text{nr}}$ with $\text{F}_{\text{txt}}$ \\
\bottomrule
\end{tabular}
\label{tab:notation}
\end{table}

\subsection{Gaussian-guided Positional Anchoring}
Recently, the explicit way~\cite{godard2017unsupervised, godard2019digging, depthanything} to interact with the 3D priors is to estimate depths from input RGB images. However, such an approach profoundly limits the advantage of 3DGS (\ie, real-time NVS) by demanding additional deep-learning capacity. 
To mitigate this, we propose a lightweight and efficient alternative: \textbf{Gaussian-guided Positional Anchoring} inspired by~\cite{shu20223dppe, liu2022petr}, which provides strong geometric clues from initialized Gaussian positions without requiring external depth supervision.

\myparagraph{Positional Anchoring leveraging Gaussian means.}
We leverage MASt3R~\cite{mast3r} to predict an initial dense 3D point map $\{\mathbf{p}_i\}_{i=1}^N$, which serves as the initialization for our Gaussian representation $\bm{\text{G}} = \{ \mu_i, \Sigma_i, \alpha_i, c_i \}$, where $\mu_i = \mathbf{p}_i$ denotes the Gaussian center. Simultaneously, a coarse camera extrinsic matrix $\text{T} = [R \mid \mathbf{t}] \in \text{SE}(3)$ is obtained per view, also from MASt3R. 
Given the current estimate of camera pose $\text{T}$ and the intrinsic matrix $\text{K} \in \mathbb{R}^{3 \times 3}$, we project each 3D point $\mathbf{p}_i \in \mathbb{R}^3$ onto the 2D image plane as below:
\begin{align}
    \tilde{\mathbf{u}}_i &= \text{K} \cdot (R \cdot \mathbf{p}_i + \mathbf{t}) \in \mathbb{R}^3, \\
    \mathbf{u}_i &= \left[ \frac{\tilde{u}_i^x}{\tilde{u}_i^z}, \frac{\tilde{u}_i^y}{\tilde{u}_i^z} \right] \in \mathbb{R}^2.
\end{align}
Each projected 2D location $\mathbf{u}_i$ serves as a spatial anchor, from which we derive a positional embedding using either sinusoidal encoding or a lightweight MLP.
\begin{equation}
\label{eq:pe}
\texttt{PE}_i = \Phi \left( \left[ \sin(\lambda^\top\mathbf{u}_i)\oplus \cos(\lambda^\top\mathbf{u}_i) \right] \right),
\end{equation}
where $\lambda$ represents a learnable frequency scale and $\oplus$ denotes concatenation.
$\Phi(\cdot)$ is a multilayer perceptron (MLP) that maps the encoded coordinates to a latent embedding space.
This formulation enables efficient and geometry-aware interaction with 3D points directly on the image plane by providing a \textit{unified positional reference}.
Crucially, it preserves spatial correspondence and depth continuity without incurring the cost of full-scale depth estimation, thus maintaining the efficiency and lightweight design of 3DGS.

\subsection{NIRSplat: A Multimodal Gaussian Splatting}
\label{subsec:NIRSplat}

\myparagraph{Bridging the invisible: NIR-RGB Coupling.}
3DGS~\cite{3dgs} introduces a powerful Gaussian-based representation that enables real-time novel view synthesis, driving substantial progress across numerous 3D vision applications.  
However, in outdoor agricultural scenarios, where sensor configurations are often sparse and viewpoint coverage is inherently limited, we observe a considerable performance drop due to inconsistent lighting, occlusion, and textureless surfaces.
To overcome these challenges, we incorporate \textbf{near-infrared (NIR)} sensing as a complementary modality to RGB.  
NIR images capture electromagnetic wavelengths beyond the visible spectrum, revealing latent structural information such as chlorophyll absorption, leaf water content, and surface reflectance properties that are often invisible in RGB.  
By leveraging this spectral prior, we aim to enhance feature robustness under adverse imaging conditions.
To this end, we design a Transformer-based NIR-RGB fusion module using a deformable cross-attention mechanism $D\_Attn$~\cite{vaswani2017attention, zhu2020deformable}.  
Let $\mathbf{F}_{rgb} = \{ f_{rgb}^i \}_{i=1}^N$ and $\mathbf{F}_{nir} = \{ f_{nir}^i \}_{i=1}^N$ be the extracted features from RGB and NIR branches.  
Each modality is augmented with a shared positional encoding $\texttt{PE}_i = \texttt{PE}[:, u_i, v_i]$, and fused via:
\begin{equation}
\label{eq:nr_attn}
\mathbf{F}_{nr}^{(i)} = D\_Attn(f_{rgb}^{(i)} \oplus \texttt{PE}_i,\ f_{nir}^{(i)} \oplus \texttt{PE}_i,\ f_{nir}^{(i)} \oplus \texttt{PE}_i)
\end{equation}
Here, $D\_Attn(\cdot)$ applies a multi-head deformable attention operation.  
This formulation allows the RGB features to selectively attend to informative NIR signals guided by spatial anchors from the shared positional encoding.
The fused representation is obtained by stacking $L$ attention layers, yielding the final robust cross-modal feature set:
$\mathbf{F}_{nr} = \left\{ \mathbf{f}_{nr}^{i} \right\}_{i=1}^{N}$, 
where $\mathbf{F}_{nr}^{(i)} \in \mathbb{R}^{C}$.
This transformer-driven NIR-RGB interaction enables effective exploitation of cross-spectral cues under limited views, leveraging both radiometric contrast from NIR and geometric alignment via positional encoding.  
As demonstrated in our experiments, this mechanism significantly enhances scene understanding and 3D reconstruction quality under real-world agricultural constraints.

\begin{table*}[t]
    \begin{center}
    \caption{\small Main Performance with SOTA techniques~\cite{fan2024instantsplat,zhang2024cor, mihajlovic2024splatfields} on \textbf{NIRPlant} dataset. We conduct experiments with 3, 6, and 12 view setups and calculate traditional three metrics: \textbf{SSIM}, \textbf{PSNR}, and \textbf{LPIPS}. 200, 1k, 10k and 30k denote iterations. Note that \textbf{bold} values indicate the best performance. \colorbox{gray!20}{Gray shading} indicates Ours.}
    \vspace{-1.0em}
    \label{tab:main}
    \resizebox{0.9\linewidth}{!}{%
    \begin{tabular}{m{3.5cm}|C{1.5cm}C{1.5cm}C{1.5cm}|C{1.5cm}C{1.5cm}C{1.5cm}|C{1.5cm}C{1.5cm}C{1.5cm}}
    \toprule
 \multirow{2.2}{*}{\textbf{Method}}   & \multicolumn{3}{c|}{\textbf{SSIM}~($\uparrow$)}   & \multicolumn{3}{c|}{\textbf{PSNR}~($\uparrow$)}  & \multicolumn{3}{c}{\textbf{LPIPS}~($\downarrow$)} \\
 \cmidrule{2-10}
   & 3-view   & 6-view  & 12-view   & 3-view   & 6-view  & 12-view     & 3-view   & 6-view  & 12-view             
\\ \midrule
3DGS                                                    
& 0.5074 & 0.5590 & 0.6531 & 14.1552 & 15.5586 & 17.4352 & 0.4586 & 0.4469 & 0.4033 \\
\cmidrule{1-10}
CoR-GS-1k 
& 0.7179 & 0.7642 & 0.8081 & 16.3494 & 17.1991 & 19.5895 & 0.4124 & 0.3191 & 0.2289 \\
CoR-GS-10k 
& 0.7285 & 0.7776 & 0.8287 & 16.7118 & 18.8023 & 20.8714 & 0.4049 & 0.3094 & 0.2281 \\
CoR-GS-30k 
& 0.7143 & 0.7611 & 0.8131 & 15.8925 & 17.5348 & 20.2927 & 0.4120 & 0.3405 & 0.2489 \\
\cmidrule{1-10}
SplatFields-1k 
& 0.7429 & 0.7647 & 0.7886 & 11.5490 & 13.6087 & 13.4497 & 0.4301 & 0.3787 & 0.3164 \\
SplatFields-10k 
& 0.7624 & 0.7799 & 0.8070 & 12.1196 & 14.6183 & 14.4463 & 0.3965 & 0.4017 & 0.2898 \\
SplatFields-30k 
& 0.7664 & 0.7802 & 0.8163 & 12.8751 & 14.1314 & 15.6037 & 0.3764 & 0.3754 & 0.2721 \\
\cmidrule{1-10}
InstantSplat-200
& 0.7559 & 0.7604 & 0.7720 & 17.6177 & 17.9250 & 18.5293 & 0.3048 & 0.2943 & 0.2784 \\
InstantSplat-1k              
& 0.7984 & 0.8126 & 0.8134 & 18.3849 & 18.9233 & 19.0333& 0.2797 & 0.2689 & 0.2438\\ 
\cmidrule{1-10}
\coloredrowcell{DCDCDC}\coloredcell{DCDCDC}NIRSplat-200
& 0.7906 & 0.8099 & 0.8174 & 18.1747 & 18.7103 & 19.1921 & 0.2371  & 0.2267 & 0.2229 \\
\coloredrowcell{DCDCDC}\coloredcell{DCDCDC}NIRSplat-1k 
& \textbf{0.8268} & \textbf{0.8311} & \textbf{0.8421} & \textbf{20.7182} & \textbf{21.0169} & \textbf{21.0814} & \textbf{0.2070} & \textbf{0.2071} & \textbf{0.2080}\\ 
\bottomrule
\end{tabular}}
\vspace{-1.0em}
\end{center}
\end{table*}

\myparagraph{Bridging the invisible: RGB-Text Coupling.}
Vision-Language Models (VLMs)~\cite{radford2021learning, alayrac2022flamingo, li2022blip, li2022grounded, zhang2023video, li2023blip, liu2023visual} have recently achieved striking success across a wide range of tasks by tightly coupling visual inputs with rich textual descriptions.
Despite their proven potential, these models remain largely unexplored in the domain of agricultural 3D reconstruction (\ie, a field that urgently demands robust, high-level scene understanding to support smart farming systems).
To address this gap, we propose a Transformer-based \textit{RGB-Text interaction module} that semantically bridges RGB features with language-derived plant attributes, enabling better recognition of hard samples (\eg, small objects, fine structures, and hard-to-perceive regions).
Primarily, we observe that various factors (\eg, descriptions, object attributes, environmental cues) in text prompts significantly contribute to view understanding~\cite{oh2024mevg, in2024editsplat, roh2024catsplat, lee2024parrot} by guiding the model's attention and perspectives. 
Inspired by this, we generate botanical-aware prompts $\mathcal{T} \in \mathbb{R}^{L \times C}$ that encapsulate detailed semantic information, such as vegetation index (\eg, NDVI, NDWI), structural traits (\eg, leaf shape, stem thickness), phenological stages (\eg, sprouting, flowering), and context (\eg, lighting, occlusion), as detailed in the supplementary document.
These prompts are first encoded using a pre-trained VLM to obtain token-level text features:
$\mathbf{F}_{txt} = \{ f_{txt}^{(1)}, f_{txt}^{(2)}, ..., f_{txt}^{(L)} \}$,
where each $f_{txt}^{(i)} \in \mathbb{R}^C$ represents a contextualized embedding.
To align language and vision, we also leverage the \textit{deformable attention} mechanism $D\_Attn$~\cite{vaswani2017attention, zhu2020deformable}, injecting shared positional priors via $\texttt{PE}$ (see Eq.~\ref{eq:pe}) in the same manner.  
We formulate the multimodal features from the NIR-RGB fusion $\mathbf{F}_{nr}^{(i)}$ and textual tokens $f_{txt}^{(i)}$ at pixel coordinate $(u_i, v_i)$ as follows:
\begin{equation}
\label{eq:rt_attn}
\mathbf{F}_{ntr}^{(i)} = D\_Attn(f_{nr}^{(i)} \oplus \texttt{PE}_i,\ f_{txt}^{(i)} \oplus \texttt{PE}_i,\ f_{txt}^{(i)} \oplus \texttt{PE}_i)
\end{equation}
Here, $\oplus$ denotes vector concatenation, and the attention module facilitates fine-grained alignment between visual and linguistic features at both spatial and semantic levels.
The resulting multimodal feature $\mathbf{F}_{\mathrm{ntr}}$ integrates geometric anchors and contextual cues and is subsequently processed by a lightweight feed-forward decoder: $\{\bm{\mu}, \bm{\alpha}, \bm{\Sigma}, \bm{c}\} = \mathrm{MLP}_{\text{gauss}}(\mathbf{F}_{ntr})$, where $\bm{\mu} \in \mathbb{R}^3$ denotes the 3D Gaussian mean, $\bm{\alpha}$ is the opacity, $\bm{\Sigma} \in \mathbb{R}^{3 \times 3}$ represents the anisotropic covariance (or its low-rank approximation), and $\bm{c}$ is the RGB appearance feature.
These learned $\bm{\text{G}}$ are directly fed into a 3D Gaussian Splatting renderer equipped with an Adaptive Density Control (ADC) mechanism, allowing efficient, robust, and botanic-aware scene reconstruction under novel agricultural scenarios (\ie, insufficient visual clues).

\subsection{Cross-Modal Gaussian Field Reasoning}

To successfully render a set of Gaussians $G$ and corresponding pose $T$, we adopt a Gaussian rasterization as a differentiable operator following Gaussian Bundle Adjustment~\cite{instantsplat} in a self-supervised manner.
Specifically, after a highly informative cross-modal fusion phase, the refined pose \textbf{T} and Gaussian field \textbf{G} are jointly optimized by minimizing the photometric rendering loss:
\begin{equation}
    \text{G}^*\text{T}^* = \text{arg}\min_{\text{G},\text{T}}\sum_{v \in N}\sum_{i=1}^{HW}\Big\Vert \tilde{C}^i_v(\text{G},\text{T})-C^i_v(\text{G},\text{T}) \Big\Vert,
\end{equation}
where $C$ and $\tilde{C}$ are the rasterization function and the observed 2D images, respectively.
Consequently, it is worth noting that this formulation facilitates rapid optimization, seamlessly incorporating complementary multimodal knowledge into the underlying 3D Gaussian representation.

\section{Experiments}
\label{sec:exp}

\begin{table*}[t!]
    \caption{\small Ablation study on various configurations.}
    \label{tab:ab1}
    \centering
    \vspace{-1.0em}
    \resizebox{0.9\linewidth}{!}{%
    \begin{tabular}{p{3cm}|l|ccc|ccc|ccc}
        \toprule
        \multirow{2.2}{*}{\textbf{Method}} & \multirow{2.2}{*}{\textbf{Configuration}} & \multicolumn{3}{c|}{\textbf{SSIM}~($\uparrow$)} & \multicolumn{3}{c|}{\textbf{PSNR}~($\uparrow$)} & \multicolumn{3}{c}{\textbf{LPIPS}~($\downarrow$)} \\
        \cmidrule{3-11}
        & & 3-view & 6-view & 12-view & 3-view & 6-view & 12-view & 3-view & 6-view & 12-view \\
        \midrule
        \multirow{6}{*}{InstantSplat-S}
        & $I_{rgb}$ only & 0.7984 & 0.8126 & 0.8134 & 18.3849 & 18.9233 & 19.0333 & 0.2797 & 0.2689 & 0.2438 \\
        & $I_{rgb}$, $I_{nir}$ & 0.7096 & 0.7383 & 0.7426 & 16.9913 & 17.6154 & 17.6345 & 0.2431 & 0.2265 & 0.2264 \\
        & $F_{rgb} \oplus F_{nir}$ & 0.8049 & 0.8079 & 0.8128 & 18.0318 & 18.7785 & 19.0927 & 0.3091 & 0.2923 & 0.2881 \\
        & $F_{rgb} \oplus F_{nir} \oplus F_{txt}$ & 0.7875 & 0.7883 & 0.7938 & 16.4174 & 17.1859 & 17.5160 & 0.2933 & 0.2742 & 0.2634 \\
        & $F_{rgb} + F_{nir}$ & 0.7605 & 0.7747 & 0.7789 & 16.1966 & 16.9488 & 17.1367  & 0.3623 & 0.3503 & 0.3505 \\
        & $F_{rgb} + F_{nir} + F_{txt}$ & 0.7514 & 0.7671 & 0.7735 & 14.6166 & 15.4207 & 16.5871 & 0.3405 & 0.3353 & 0.3246 \\
        \midrule
        \multirow{3}{*}{NIRSplat-S}
        & $attn(F_{rgb},F_{txt})$ & 0.8053 & 0.8139 & 0.8160 & 18.8696 & 18.7132 & 19.1866 & 0.2765 & 0.2728 & 0.2457 \\
        & $attn(F_{rgb},F_{nir})$ & 0.8205 & 0.8240 & 0.8314 & 20.0486 & 20.2083 & 20.9963 & 0.2244 & 0.2153 & 0.2130 \\
        \coloredrowcell{DCDCDC} & $attn(F_{rgb},F_{nir},F_{txt})$ & \textbf{0.8268} & \textbf{0.8311} & \textbf{0.8421} & \textbf{20.7182} & \textbf{21.0169} & \textbf{21.0814} & \textbf{0.2070} & \textbf{0.2071} & \textbf{0.2080}\\ 
        \bottomrule
    \end{tabular}
    }
    \vspace{-1.0em}
\end{table*}

\begin{figure*}[t]
    \centering
    \includegraphics[width=0.9\linewidth]{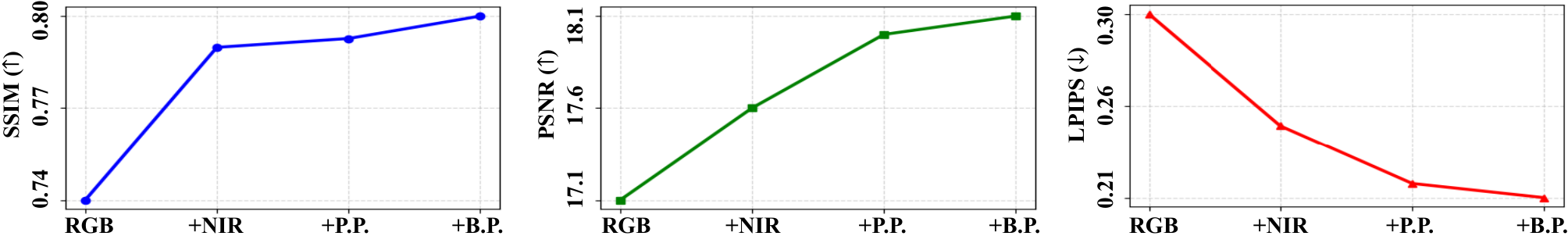}
    \vspace{-0.4em}
    \caption{\small Comparison under challenging lighting conditions. P.P. and B.P. denote plain prompts and botanic-aware prompts.}
    \label{fig:light}
\end{figure*}

\myparagraph{Baselines.} 
We selected recent state-of-the-art methods for comparison, including 3DGS~\cite{3dgs}, SplatFields~\cite{mihajlovic2024splatfields}, InstantSplat~\cite{fan2024instantsplat} and CoR-GS~\cite{zhang2024cor}. 
These methods efficiently leverage a Gaussian parameter in real time, optimizing a position $\bm{\mu}$, an opacity $\bm{\alpha}$, a covariance $\bm{\Sigma} \in \mathbb{R}^{3 \times 3}$, and spherical harmonics (color) $c$ with trivial computational overhead. 
Additionally, we adopt pose-free methods, Nope-NeRF~\cite{bian2023nope} and CF-3DGS~\cite{fu2024colmap}, which are supported by monocular depth maps and ground-truth camera intrinsics, following InstantSplat.
Please refer to the supplementary document for detailed experimental setups.

\subsection{Benchmarks and Discussions}
\label{sec:benchmark}

Previous works often struggle under agricultural conditions due to uncertain camera perspectives, insufficient visual cues, and constraints on computational resources, making robust 3D reconstruction particularly challenging.
Specifically, 3DGS suffers from limited 2D information (\ie, due to sparse-view setups), leading to an inevitable performance drop (up to -31.9$\%$ SSIM, -5.8 PSNR, and +25.2$\%$ LPIPS gaps compared to our \textbf{NIRSplat}), as shown in Tab.~\ref{tab:main}. 
In the 3-view configuration, CoR-GS shows substantial structural degradation (lowest SSIM score) among recent sparse-view approaches~\cite{mihajlovic2024splatfields, fan2024instantsplat}. 
Meanwhile, SplatFields fails to preserve pixel-level fidelity, resulting in a notable drop in PSNR and suboptimal reconstruction quality.
Furthermore, these models exhibit poor performance under extremely limited training budgets (1k iterations), suggesting a lack of inherent robustness.
Although InstantSplat addresses these drawbacks, this paradigm is still limited in capturing visual details from challenging agricultural samples (\ie, occlusion, uneven reflection), resulting in up to -2.6$\%$ SSIM, -2.5 PSNR, and +7.2$\%$ LPIPS loss, compared to Ours.
To tackle these issues, we leverage a novel multi-modal architecture, \textbf{NIRSplat}, which effectively generalizes agricultural environments. Notably, \textbf{NIRSplat} demonstrates its efficiency and validity by surpassing the performance of previous models that use 12 views despite using only 3 views.

\begin{figure*}[t]
    \centering
    \includegraphics[width=0.9\linewidth]{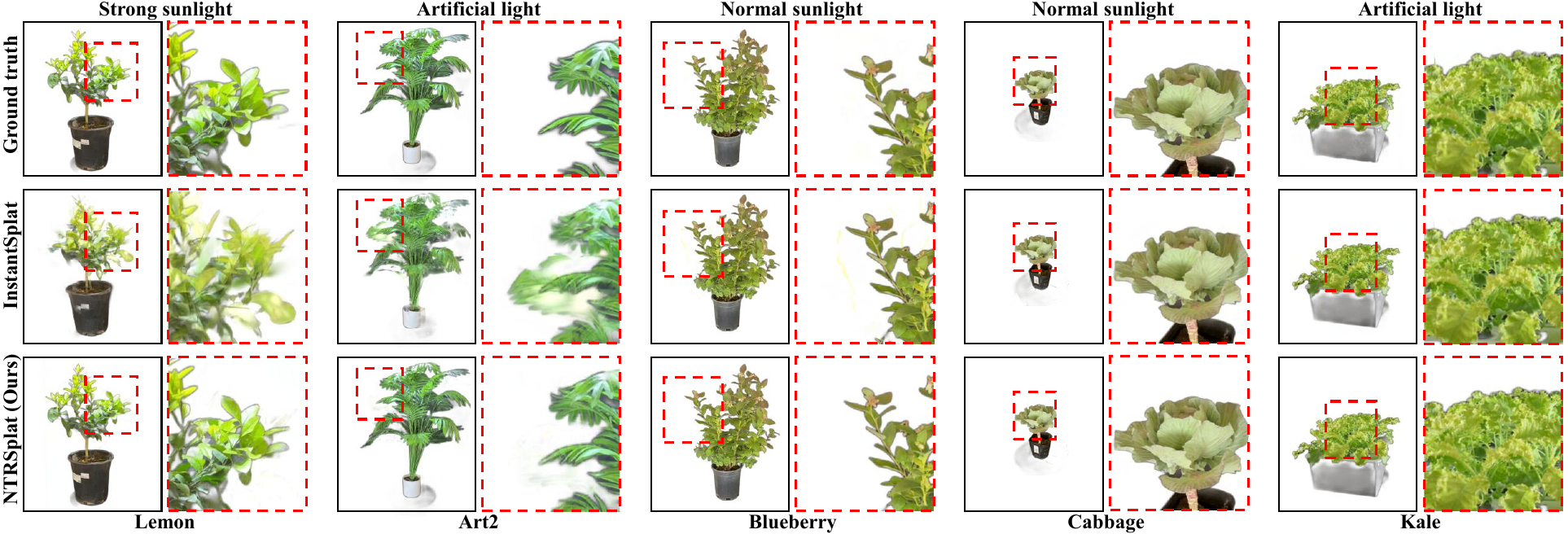}
    \vspace{-0.6em}
    \caption{\small Qualitative visualization in a 3-view setup, demonstrating the results under diverse lighting conditions: Lemon (strong light), Kale (artificial light), Art2 (occlusion), and Cabbage (small object). The red box highlights the semantic loss associated with conventional methods.}
    \label{fig:qualitative}
    \vspace{-0.8em}
\end{figure*}


\begin{table}[t]
    \begin{center}
    \caption{Ablation study of \texttt{PE} (\cref{eq:pe}).}
    \vspace{-0.8em}
    \label{tab:pe}
    \resizebox{0.8\linewidth}{!}{%
    \begin{tabular}{C{1.2cm}C{1.2cm}|C{1.6cm}C{1.6cm}C{1.6cm}}
    \toprule
 \multicolumn{2}{c}{\textbf{\texttt{PE}}} & \multicolumn{3}{|c}{\textbf{6 views}} \\
 \cmidrule{1-5}
 w/o & w/ & \textbf{SSIM~($\uparrow$)} & \textbf{PSNR~($\uparrow$)} & \textbf{LPIPS~($\downarrow$)} \\
 \midrule
 \checkmark & & 
 0.8159 & 18.3132 & 0.2745 \\
 \coloredrowcell{DCDCDC}\coloredcell{DCDCDC} 
 & \checkmark & 
 \textbf{0.8311} & \textbf{21.0169} & \textbf{0.2071} \\
\bottomrule
\end{tabular}}
\vspace{-1.4em}
\end{center}
\end{table}


\subsection{Ablation Studies}
\label{sec:ablation}
\myparagraph{Impact of Additional Modalities.} 
This naturally raises a fundamental question: \textit{Do additional modalities consistently yield performance improvements?}
While additional modalities (NIR, Text) provide rich complementary cues, seamlessly integrating them remains a significant challenge. 
This difficulty largely stems from inherent modality gaps, spectral discrepancies, and disjoint embedding spaces across RGB, NIR, and textual inputs.
To better understand this, we explore various fusion strategies in~\cref{tab:ab1}.
Naïvely adding NIR signals significantly degrades performance, leading to up to a \textbf{9\% drop in SSIM}. We attribute this to the \textit{spectral discrepancy} and value distribution mismatch between visible and near-infrared modalities.
Conventional fusion techniques (element-wise summation, feature concatenation) result in trivial improvements, failing to resolve the semantic and spatial misalignment.
Importantly, this limitation potentially becomes more pronounced when incorporating textual metadata as shown in~\cref{tab:ab1} and~\cref{tab:pe}: without proper geometric references, semantic and geometric misalignment between visual and textual features causes suboptimal training (\ie, \textbf{-1.52\% SSIM}, \textbf{-2.7 PSNR}, \textbf{+6.74\% LPIPS}).
To address this, we introduce an effective cross-modal 3D reconstruction method, \textbf{NIRSplat} that leverages a geometry-guided 3D point-based positional encoding (\texttt{PE}) scheme anchoring features from all modalities to a shared 2D projection space.
Consequently, NIRSplat facilitates robust alignment for uncertain cross-modal knowledge by allowing the model to leverage the most informative signals from each modality, thereby achieving superior 3D consistency and fidelity.

\myparagraph{Effect of Botanic-Aware Knowledge.}
To further enhance scene understanding in agricultural domains, we incorporate botanic-aware textual prompts that encode physiological and environmental context (\ie, NDVI, NDWI, chlorophyll levels, growth stages, and lighting conditions), as detailed in the supplement.
One might reasonably question the effectiveness of language-based guidance in dense 3D reconstruction, as textual descriptions are often semantically abstract and may lack spatial precision.
To address this concern, we conduct an ablation study comparing plain prompts with botanic-aware prompts that explicitly embed spectral and biological indices.
As shown in~\cref{fig:light}, botanic-aware prompts lead to consistent performance gains, with improvements compared to plain prompts under challenging scenarios.
These results indicate that botanic-aware prompts, unlike plain prompts, act as high-level priors that reinforce correlations between NIR responses and botanical states, guiding cross-modal attention toward semantically and structurally relevant regions and improving reconstruction fidelity under ambiguity or occlusion.

\subsection{Qualitative Analyses}
We qualitatively evaluate our method under four challenging agricultural scenarios: (i) \textbf{Strong sunlight} (Lemon), (ii) \textbf{artificial lighting} (Kale, Art2), (iii) \textbf{occlusion} (Art2), and (iv) \textbf{small objects} (Cabbage), with Blueberry serving as a moderate-complexity reference (see~\cref{fig:qualitative}).
Conventional methods (\eg, InstantSplat) often fail to preserve semantic and geometric fidelity, showing blurred textures and structural collapse under occlusion or extreme lighting. In contrast, \textbf{NIRSplat} achieves clear improvements by leveraging spectral cues (e.g., NIR) and botanic-aware priors, enabling better detail recovery and structural consistency.
Notably, NIRSplat recovers saturated regions under strong illumination, resolves fine details in small-scale objects, and maintains coherence in occluded or low-texture areas—demonstrating its robustness across diverse agricultural conditions.

\section{Conclusion}
\label{sec:conclusion}

\myparagraph{Summary.} 
In this work, we introduced the \textbf{NIRPlant} dataset, which incorporates multimodal data from Near-Infrared (NIR), text, and RGB sensors in both indoor and outdoor agricultural environments. By leveraging the unique advantages of NIR and botanical-aware text, we addressed the challenges of 3D reconstruction in agriculture, including uneven lighting, occlusion, and novel perspectives. We also presented \textbf{NIRSplat}, an effective multimodal Gaussian Splatting framework that bridges these modalities through cross-attention and strong geometric priors from 3D point-based positional encoding. Importantly, \textbf{NIRSplat} significantly improves scene understanding, leveraging invisible NIR and contextual text knowledge. Through comprehensive experiments, we demonstrated that \textbf{NIRSplat} outperforms state-of-the-art methods, highlighting the potential of multimodal integration for robust agricultural 3D reconstruction.
 
\myparagraph{Limitations and Future Work.} 
We found that additional inputs lead to significant computational overheads, which limit the efficiency of real-time rendering. While our transformer-based approach effectively bridges the multimodality, it suffers from the cost of increased model complexity and capacity. 
In future work, we aim to address this issue by seamlessly aligning the three different modalities, ensuring more efficient integration and reducing computational overhead.

\bibliography{main}

\clearpage
\setcounter{page}{1}
\maketitlesupplementary


\section{Motivation and Practical Value} 
Incorporating additional modalities (\eg, NIR, LiDAR, and botanical metadata) poses challenges, as they may not always be feasible in general-purpose applications. However, our focus is not on general-purpose 3D reconstruction but on automated smart farming—an area where such infrastructure is already being adopted. In commercial precision agriculture, multi-spectral cameras and LiDAR sensors are now standard, with associated metadata (\eg, NDVI, NDWI) often computed automatically by embedded software. In scenarios where high-fidelity reconstruction and crop monitoring are critical (\ie, phenotyping facilities), we argue that using these modalities is both practical and justified, especially compared to the cost of inaccurate measurements. To be clear, we are not advocating for universal deployment of all modalities, but rather demonstrating that, when such rich data is available, our framework can effectively leverage it to achieve robust and accurate reconstruction under challenging conditions (\eg, occlusions, uneven lighting). It is noteworthy that as agricultural technology advances, multimodal sensing is becoming increasingly accessible, and methods that can capitalize on it will be essential for future progress.


\section{Additional Related Works}
\subsection{Multimodal 3D Reconstruction}
Multimodal 3D Reconstruction methods leverage the complementary strengths of multiple data types to produce more accurate reconstructions. Traditional multimodal models that combine RGB image data and depth data have commonly been used to produce dense, high-quality reconstructions of 3D surfaces~\cite{izadi2011kinectfusion, zollhofer2018state}. The use of other data types, including infrared, time-of-flight, white light interferometry, and cone beam computed tomography data, for multimodal 3D Reconstruction has also been explored~\cite{liu2023multi, stotko2019albedo, chen2025opticfusion,hao2022ai}.

Recently, neural network-based multimodal 3D reconstruction methods have been developed. Differentiable volume rendering techniques from single-modal Reconstruction have been applied to multimodal 3D reconstruction~\cite{liu2023multi,sun2024mm3dgs}. Modern data encoders such as BERT and CLIP allow 3D scene completion and generation guided by a combination of text, image, and incomplete scene data~\cite{cheng2023sdfusion}.  

\section{Preliminary}
\label{sup:pre}

\myparagraph{3D Gaussian Splatting.} We build upon the foundational 3D Gaussian Splatting (3DGS) method~\cite{3dgs}, which represents a 3D scene using a set of anisotropic Gaussian primitives. As shown in~\cref{eq:3dgs}, each primitive is parameterized by a 3D position $\bm{\mu} \in \mathbb{R}^3$, an opacity value $\alpha$, a full covariance matrix $\bm{\Sigma} \in \mathbb{R}^{3 \times 3}$ encoding spatial uncertainty, and view-dependent color modeled via spherical harmonics (SH) as $\bm{c}(\textbf{d}) = \sum_i c_i \mathcal{B}_i(\textbf{d})$, where $\mathcal{B}_i$ are the SH basis functions.
\begin{equation}
    \label{eq:3dgs}
    \bm{\text{G}}(\textbf{\text{p}},\alpha,\bm{\Sigma}) = \alpha~\text{exp}(-\frac{1}{2}(\textbf{\text{p}}-\textbf{\text{x}})^T\Sigma^{-1}(\textbf{\text{p}}-\mu)).
\end{equation}
Unlike volumetric radiance fields (\eg, NeRF) or voxel-based approaches, 3DGS eliminates the need for expensive ray marching by directly projecting the Gaussians into screen space, enabling efficient rasterization and differentiable rendering. This results in significantly improved rendering speed, compact representation, and high fidelity for real-time novel view synthesis, making it highly suitable for dynamic or resource-constrained applications.


\begin{table*}[t!]
\centering
\caption{Botanic-aware Prompt Engineering.}
\label{tab:prompt}
\resizebox{1.0\linewidth}{!}{%
\begin{tabular}{m{20cm}}
\toprule
\textbf{Botanic-aware Prompt Engineering} \\
\midrule
You are a multimodal plant understanding expert specializing in 3D reconstruction under challenging visual and environmental conditions.
Your task is to generate a single coherent paragraph by synthesizing (i) image description, (ii) NIR-based indices (NDVI, NDWI, chlorophyll index), and (iii) weather metadata, highlighting structural and physiological features to support cross-modal attention in downstream reconstruction. \\
\\
The provided data are as follows:

\textbf{Image Description}: \{\texttt{img\_description}\} \\

\textbf{NIR Information}: \{\texttt{NDVI, NDWI, chlorophyll index}\} \\

\textbf{Weather Information}: \{\texttt{temp, dew, humidity, precip, precipprob, cloudcover, solarradiation, uvindex, windgust, windspeed, visibility}\}\\

\\
\textbf{Instructions:}
\begin{itemize}
    \item Do not simply concatenate the inputs. Instead, synthesize them into a natural and cohesive narrative.
    \item Evaluate whether the weather conditions and image description are semantically aligned (e.g., indoor/outdoor consistency). Use only the relevant components.
    \item Embed physiological and environmental data meaningfully rather than listing raw numbers (e.g., say "under high humidity" instead of "81\%").
    \item Emphasize attributes that affect visual appearance and 3D geometry, such as occlusion, lighting variation, leaf texture, or overlapping structures.
    \item Limit your output to a paragraph (max 200 words) in fluent and formal English.
    \item Do not include explanations, metadata, or any extra text. Output only the final paragraph.
\end{itemize}
\\
\textbf{The final rewritten description is:} \\
\bottomrule
\end{tabular}%
}
\end{table*}


\myparagraph{Multi-View Stereo.} 
MASt3R incorporates a dense local feature head and a fast reciprocal nearest-neighbor matching scheme and optimizes the pixel-aligned point maps $\textbf{P}$ directly from raw images. 
Unlike traditional 2D matching approaches, MASt3R treats matching as an intrinsically 3D task by regressing point maps in a shared coordinate frame, thus enabling robust correspondence even under extreme viewpoint variations.
In this paper, building on the feed-forward Multi-View Stereo framework MASt3R~\cite{mast3r}, we introduce a tri-modal architecture NIRSplat that takes as input aligned RGB frames $\mathcal{I}_{rgb}^{N} \in \mathbb{R}^{N\times H\times W\times3}$, single-channel NIR frames $\mathcal{I}_{nir}^{N} \in \mathbb{R}^{N\times H\times W\times1}$, and contextual text prompts $\mathcal{T} \in \mathbb{R}^{L\times C}$.
Dedicated encoders transform each modality into latent sequences; $\mathbf{F}_{rgb}=\{f_{rgb}^i\}_{i=1}^{N}$, $\mathbf{F}_{nir}=\{f_{nir}^i\}_{i=1}^{N}$, and $\mathbf{F}_{txt}=\{f_{txt}^i\}_{i=1}^{L}$ are subsequently fused by a deformable-cross-attention Transformer \cite{vaswani2017attention,zhu2020deformable}.
This fusion module aligns spectral (RGB $\leftrightarrow$ NIR) and semantic (vision $\leftrightarrow$ language) cues in a shared token space, letting attention weights act as adaptive modality gates: noisy RGB pixels under adverse lighting conditions are down-weighted, while light-robust NIR edges and text-driven botanical priors are amplified.


\myparagraph{Loss function.}
Motivated by recent studies~\cite{fan2024instantsplat, mast3r_arxiv24, dust3r_cvpr24}, we directly penalize between the predicted point maps $\mathbf{P}$ and the pseudo ground truths $\hat{\mathbf{P}}$ from MASt3R using the following regression loss term:
\begin{equation}
    \label{reg_loss}
    \mathcal{L}_{reg}=\Big\Vert \frac{1}{z_i} \cdot \mathbf{P}_{v,1} - \frac{1}{z_i} \cdot \hat{\mathbf{P}}_{v,1} \Big\Vert,
\end{equation}
where $v \in \{1,2\}$, $z$ denote corresponding views and normalization factors. 
The predicted Gaussian means $\mathbf{P}_{v,1}$ are obtained via our NIRSplat framework, which enables efficient semantic grounding in 3D space.
Plus, we optimize the pixel-aligned confidence score $O^i_{v,1}$ to improve scene understanding through the following objective:
\begin{equation}
    \mathcal{L}_{conf} = \sum_{v \in {1,2}}{\sum_{i \in D^v}}{O^i_{v,1} \cdot \mathcal{L}(v,i)-\alpha \cdot \text{log}O^i_{v,1}}.
\end{equation}
where $\mathcal{L}(v,i)$ denotes \cref{reg_loss}, and $\alpha$ is a balancing weight.
The first term encourages accurate predictions in confident regions, while the second regularizes overconfidence via a log penalty.
This formulation enables the model to selectively focus on spatially reliable regions and suppress uncertain noise, resulting in substantial improvements in both semantic fidelity and geometric precision.


\section{Botanic-aware Prompt Engineering}

To facilitate robust scene understanding under challenging agricultural conditions, we design a botanic-aware prompt engineering strategy (\cref{tab:prompt}) that synthesizes complementary visual, physiological, and environmental modalities into a unified textual representation. Our prompt integrates (i) image descriptions highlighting structural features such as occlusion, overlapping leaves, or lighting variation, (ii) NIR-derived physiological cues including NDVI, NDWI, and chlorophyll index, and (iii) weather conditions such as humidity, solar radiation, or wind gusts that may influence plant appearance and geometry.
Unlike naive concatenation of metadata, the prompt generation process encourages the creation of a fluent paragraph that semantically aligns these modalities, enabling the model to infer cross-modal relationships and guide attention toward informative regions. In particular, collaborating with NIR signals (\ie, highly sensitive to internal plant states) offers deep contextual cues that enrich understanding of vegetation health, moisture levels, and pigment concentration. Importantly, this multimodal prompt not only anchors spatial features but also enhances generalizability across diverse environments, contributing to downstream 3D reconstruction tasks.

\section{Experimental Configurations.}

\noindent\textbf{Implementation Detail.} 
To ensure a fair comparison with our baseline, we adopt a sparse-view 3D reconstruction paradigm using 3, 6, and 12 input views. For each setting, we perform 200 and 1,000 iterations for training and 500 iterations for test-view optimization, respectively, maintaining consistency across all experiments. Furthermore, to account for the relatively slower convergence of certain baselines, we extend the training schedules of 3DGS~\cite{3dgs}, CoR-GS~\cite{zhang2024cor} and SplatFields~\cite{mihajlovic2024splatfields} to 10k and 30k iterations, respectively, thereby providing sufficient optimization steps and mitigating potential underfitting in the sparse-view scenario. Unless otherwise stated, all methods are trained and evaluated using two NVIDIA RTX 6000 ADA GPUs under identical computational conditions.

\noindent\textbf{Evaluation Metrics.} 
For quantitative evaluation, we employ three widely adopted metrics for novel view synthesis, following prior work~\cite{3dgs,instantsplat}: 
Peak Signal-to-Noise Ratio (PSNR), Structural Similarity Index Measure (SSIM)~\cite{wang2004image}, and Learned Perceptual Image Patch Similarity (LPIPS)~\cite{zhang2018unreasonable}. 
Among these, PSNR measures the pixel-wise reconstruction fidelity, SSIM evaluates the perceived structural similarity between the reconstructed and reference images, and LPIPS assesses perceptual quality using deep feature embeddings. 
To ensure a consistent and fair comparison under sparse-view conditions, we strictly follow the evaluation protocol of the state-of-the-art InstantSplat~\cite{instantsplat}, including its sparse-view reconstruction settings and test-view sampling strategy.

\begin{table*}[t]
    \begin{center}
    \caption{Ablation study on \textbf{NIRPlant}. Text, NIR, and Temp represent RGB-Text Interaction, NIR-RGB Interaction, and Temporal Interaction, respectively. We conduct an ablation study with 3, 6, and 12 view setups and 1000 iterations. Note that \textbf{bold} values indicate the best performance.}
    \label{tab:ab2}
    \resizebox{\linewidth}{!}{%
    \begin{tabular}{C{1cm}C{1cm}|C{1.5cm}C{1.5cm}C{1.5cm}|C{1.5cm}C{1.5cm}C{1.5cm}|C{1.5cm}C{1.5cm}C{1.5cm}}
    \toprule
 \multicolumn{2}{c|}{\textbf{Method}} & \multicolumn{3}{c|}{\textbf{SSIM}~($\uparrow$)}   & \multicolumn{3}{c|}{\textbf{PSNR}~($\uparrow$)}  & \multicolumn{3}{c}{\textbf{LPIPS}~($\downarrow$)}            \\
 \cmidrule{1-11}
   Text & NIR & 3-view   & 6-view  & 12-view   & 3-view   & 6-view  & 12-view     & 3-view   & 6-view  & 12-view             
\\ \hline
&& 
0.7984 & 0.8126 & 0.8134 & 18.3849 & 18.9233 & 19.0333& 0.2797 & 0.2689 & 0.2438 \\ 
                                            
\checkmark & &  
0.8053 & 0.8139 & 0.8160 & 18.8696 & 18.7132 & 19.1866 & 0.2765 & 0.2728 & 0.2457 \\
& \checkmark &  
0.8205 & 0.8240 & 0.8314 & 20.0486 & 20.2083 & 20.9963 & 0.2244 & 0.2153 & 0.2130 \\

\coloredrowcell{DCDCDC}\coloredcell{DCDCDC}
\checkmark&\checkmark&
\textbf{0.8268} & \textbf{0.8311} & \textbf{0.8421} & \textbf{20.7182} & \textbf{21.0169} & \textbf{21.0814} & \textbf{0.2070} & \textbf{0.2071} & \textbf{0.2080}\\ 

\bottomrule
\end{tabular}}
\end{center}
\end{table*}


\begin{table}[t]
    \begin{center}
    \caption{Ablation study of metadata. We train our model with six view setups and 1000 iterations. Note that \textbf{bold} values indicate the best performance.}
    \label{tab:meta}
    \resizebox{\linewidth}{!}{%
    \begin{tabular}{C{1.5cm}C{1.5cm}|C{1.7cm}C{1.7cm}C{1.7cm}}
    \toprule
 \multicolumn{2}{c}{\textbf{Metadata}} & \multicolumn{3}{|c}{\textbf{6 views}} \\
 \cmidrule{1-5}
 w/o & w/ & \textbf{SSIM~($\uparrow$)} & \textbf{PSNR~($\uparrow$)} & \textbf{LPIPS~($\downarrow$)} \\
 \midrule
 \checkmark & & 0.8299 & 20.8694 & 0.2172 \\
 \coloredrowcell{DCDCDC}\coloredcell{DCDCDC} 
 & \checkmark & \textbf{0.8311} & \textbf{21.0169} & \textbf{0.2071} \\
\bottomrule
\end{tabular}}
\end{center}
\end{table}


\section{Additional Experiments.}

\myparagraph{Ablation of modalities.}
We ablate our proposed method, \textbf{NIRSplat}, to validate the contribution of each knowledge source (\ie, NIR and Text) under different view configurations (\ie, 3, 6, and 12 views) with 1000 iterations, as shown in~\cref{tab:ab2}.
First, incorporating only the text modality already leads to notable improvements over the baseline InstantSplat across all metrics. This gain is attributed to our botanic-aware textual guidance, which enhances scene understanding through weighted attention and semantically rich alignment.
Second, the inclusion of the NIR modality also contributes positively, particularly in low-visibility and occlusion-heavy conditions commonly found in agricultural environments. It yields improved perceptual quality and spatial consistency across all views.
Most importantly, when both modalities (\ie, text and NIR) are combined, the model achieves the best overall performance:
$+2.8\%$ SSIM, $+1.7\%$ PSNR, and $-5.5\%$ LPIPS improvement over the baseline at 3-view, with consistent gains across all view settings. This highlights the synergistic effect of multimodal integration in our framework.
Overall, \textbf{NIRSplat} effectively leverages complementary multimodal cues to overcome the limitations of conventional methods, demonstrating its superiority in robustness, accuracy, and perceptual fidelity.


\myparagraph{Delving into botanic-aware prompts.}
We observed that in~\cref{tab:ab2} RGB-only methods struggle to reconstruct fine-grained agricultural structures under challenging lighting conditions, mainly due to overexposure and low illumination. 
Here, we present an in-depth analysis of how botanic-aware knowledge significantly enhances scene understanding in novel agricultural environments.
First, the NIR modality contributes significantly by operating in a distinct spectral range, enabling consistent structural recovery even under harsh lighting. This yields notable gains across all metrics (see +NIR in~\cref{tab:ab2}).
Crucially, we further boost the effectiveness of text guidance through \textbf{botanic-aware prompt engineering}, where prompts explicitly encode spectral and environmental metadata (\eg, NDVI, lighting conditions). As validated in~\cref{tab:meta}, incorporating metadata yields measurable improvements: $+0.12$ SSIM, $+0.15$ PSNR, and $-0.0051$ LPIPS, demonstrating that environmental cues are not only semantically meaningful but also practically beneficial.
Furthermore, our method outperforms landmark vision-language models such as CLIP and BLIP in multi-view reconstruction tasks (see~\cref{tab:vlm}). Notably, \textbf{BLIP-2 + Metadata} achieves the best overall performance across all metrics, indicating the synergy between domain-specific prompting and adaptive attention.
During weighted cross-attention, NIRSplat dynamically assigns higher attention to prompt tokens enriched by these metadata-driven priors, thereby enhancing modality alignment and scene-level reasoning. This selective focus enables the model to effectively decode and integrate $\mathcal{F}_{NR}$—the fused feature from NIR and RGB—resulting in perceptually faithful and structurally accurate reconstructions.


\begin{table}[t]
    \begin{center}
    \caption{Comparison with landmark VLMs. We train our model with six view setups and 1000 iterations. Note that \textbf{bold} values indicate the best performance.}
    \label{tab:vlm}
    \resizebox{\linewidth}{!}{%
    \begin{tabular}{c|ccc}
    \toprule
 \multirow{2.2}{*}{\textbf{Method}} & \multicolumn{3}{c}{\textbf{6 views}} \\
 \cmidrule{2-4}
 & \textbf{SSIM~($\uparrow$)} & \textbf{PSNR~($\uparrow$)} & \textbf{LPIPS~($\downarrow$)} \\
 \midrule
CLIP~\cite{radford2021learning}   & 0.8281 & 20.2083 & 0.2080\\
BLIP~\cite{li2022blip}   & 0.8208 & \textbf{21.0330} & 0.2090 \\
\coloredrowcell{DCDCDC}\coloredcell{DCDCDC}
BLIP-2~\cite{li2023blip} & \textbf{0.8311} & 21.0169 & \textbf{0.2071} \\
\bottomrule
\end{tabular}}
\end{center}
\end{table}


\subsection{Additional Qualitative Analyses}

\begin{figure}[t]
    \centering
    \includegraphics[width=1\linewidth]{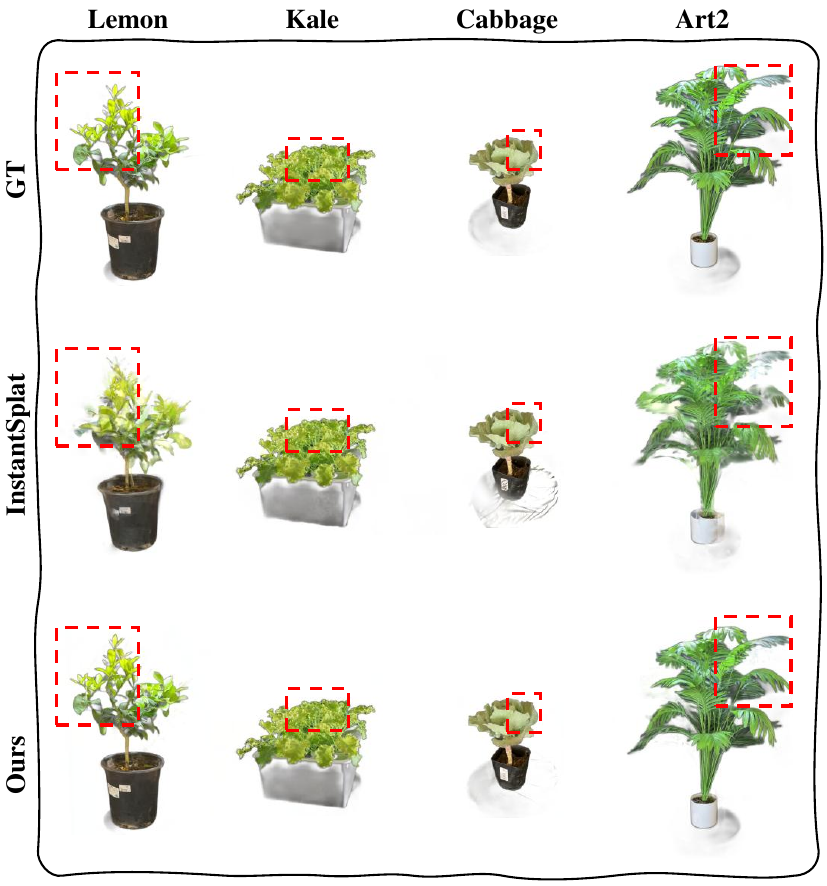}
    \caption{Qualitative visualization in a 3-view setup, demonstrating the results under diverse lighting conditions: Lemon (strong light), Kale (artificial light), Art2 (occlusion), and Cabbage (small object). The comparison highlights the semantic loss when using conventional methods, with a marked improvement achieved by our novel approach, \textbf{NIRSplat}. Zoom in for better visualization.}
    \label{supp_fig:qualitative}
\end{figure}

In this section, we qualitatively demonstrate that our proposed approach effectively addresses the core challenges of agricultural scene reconstruction by leveraging \textit{invisible} yet informative multimodal cues, as shown in~\cref{supp_fig:qualitative}. 
Specifically, we introduce four representative and challenging scenarios in which existing methods consistently fail to preserve key visual properties such as texture, shape, and volume, primarily due to ambiguous or missing semantic clues (highlighted in red dotted boxes).
Agricultural environments inherently present unique obstacles, including incomplete or skewed viewpoints, extreme lighting conditions (\eg, overexposure or underexposure), and heavy occlusion from overlapping foliage or dense vegetation. These factors often result in \textit{non-trivial} failures during 3D reconstruction, as illustrated in Fig.~\ref{supp_fig:qualitative}.
First, we observe that existing methods are highly sensitive to uncontrolled lighting (particularly in cases of whiteout caused by strong sunlight), where they tend to over-smooth or completely wash out structural content.
Second, excessive occlusion often introduces redundancy and ambiguity in depth reasoning, leading to artifacts or duplicated geometries (see the red box in Art2).
Third, reconstructing fine-scale texture and geometry—such as small and dense crops like cabbages—remains a significant challenge for prior approaches due to their limited spatial resolution and inability to preserve fine-grained surface detail, often resulting in blurry or semantically inconsistent outputs.
To overcome these limitations, we propose \textbf{NIRPlant} and our full model \textbf{NIRSplat}, which jointly incorporate NIR and text-guided prompts to adaptively handle challenging conditions. Our method robustly compensates for missing semantic cues through modality-aware attention and is able to synthesize coherent, high-fidelity 3D structures across diverse lighting, occlusion, and viewpoint variations. These results highlight the practical benefits of multimodal fusion in advancing scene understanding in real-world agricultural scenarios.


\section{Licenses} Our collected data is under the CC-BY-4.0 (\url{https://creativecommons.org/licenses/by/4.0/legalcode.en}) license. In addition, the dataset shall be used only for non-commercial research and educational purposes.

\section{Acknowledgements}
This research was supported by Culture, Sports and Tourism R$\&$D Program through the Korea Creative Content Agency grant funded by the Ministry of Culture, Sports and Tourism (Project Name: International Collaborative Research and Global Talent Development for the Development of Copyright Management and Protection Technologies for Generative AI, Project Number: RS-2024-00345025) and the National Institute of Food and Agriculture of the US Department of Agriculture (grant number 2024-67021-42528).

\end{document}